\documentclass[letterpaper]{article}

\usepackage[draft]{aaai2026} % Authors shown; copyright suppressed
\usepackage{times}
\usepackage{helvet}
\usepackage{courier}
\usepackage[hyphens]{url}
\usepackage{graphicx}
\usepackage{natbib}
\usepackage{caption}
\usepackage{algorithm}
\usepackage{algorithmic}
\usepackage{amsmath}
\usepackage{amssymb}
\usepackage{booktabs}
\usepackage{multirow}
\usepackage{xcolor}

\title{TREA-Net: A Transferable Residual Epidemiological Adaptation
Network for Dengue Incidence Forecasting}

\author{
Inesh Shukla\textsuperscript{\rm 1},
Madhurima Panja\textsuperscript{\rm 2},
Tanujit Chakraborty\textsuperscript{\rm 2,3},
Chittaranjan Hens\textsuperscript{\rm 1}
}

\affiliations{
\textsuperscript{\rm 1}
International Institute of Information Technology Hyderabad, India\\
\textsuperscript{\rm 2}
SAFIR, Sorbonne University Abu Dhabi, United Arab Emirates\\
\textsuperscript{\rm 3}
Sorbonne Center for Artificial Intelligence,
Sorbonne Université, Paris, France\\
inesh.shukla@research.iiit.ac.in,
chittaranjan.hens@iiit.ac.in,
madhurima.panja@sorbonne.ae,
tanujit.chakraborty@sorbonne.ae
}

\begin{document}
\maketitle

\begin{abstract}
Accurate multi-week dengue forecasting supports timely vector-control interventions, outbreak preparedness, and healthcare resource allocation. However, newly established surveillance systems often lack the historical data needed to train reliable neural forecasting models. Although pretrained time-series models offer promising zero-shot forecasts, their cross-domain training may not capture local epidemiological dynamics. We propose TREA-Net, a Transferable Residual Epidemiological Adaptation Network for dengue forecasting under limited data. TREA-Net augments neural forecasting backbones with projections from an Environmental Time-Series Susceptible–Infected–Recovered model and learns a lightweight gated residual correction transferable from data-rich to data-scarce regions. Its node-invariant design accommodates surveillance systems with different numbers of locations, while target adaptation requires learning only two global parameters. We transfer knowledge from long-running dengue surveillance in Colombia and Nicaragua to 8-week-ahead forecasting in Mexico and Malaysia using only 78 or 104 weeks of target data. Across five neural backbones and ten transfer settings, TREA-Net improves the corresponding backbone in 9 out of 10 settings, with statistically significant gains. When integrated with TiRex (a foundation model for forecasting), it achieves the lowest mean absolute error across all target datasets. Conformal prediction further maintains empirical coverage while reducing 8-week prediction-interval width by 29.6\% in Mexico. These results demonstrate TREA-Net’s potential as a lightweight and portable early-warning framework for health agencies with limited surveillance data.

\end{abstract}

\begin{links}
    \link{Code}{anonymous.4open.science/r/TREA-Net-C8F8/}
\end{links}

% ============================================================
\section{Introduction}

Global dengue incidence more than doubled between 1990 and 2021, reaching approximately 60 million cases annually and accounting for nearly 2 million disability-adjusted life-years worldwide \citep{gbd2021dengue, who2023dengue}. This burden is expected to increase as climate change expands the geographic range and transmission season of \textit{Aedes aegypti}, exposing regions with limited surveillance capacity to greater outbreak risk \citep{ipcc2023ar6}. Reliable multi-week forecasting is therefore increasingly important for vector-control planning, healthcare resource allocation, and public-health decision-making. Forecasts up to eight weeks ahead can provide sufficient lead time for intervention while remaining informative about near-term transmission dynamics \citep{chen2018singapore, johansson2019dengue, chakraborty2019forecasting}. Yet this task is particularly difficult for newly established surveillance systems, where short historical records limit the ability of data-driven models to learn stable temporal and spatial patterns for multivariate dengue incidence datasets.

\begin{figure}[!t]
  \centering
  \includegraphics[width=\columnwidth]{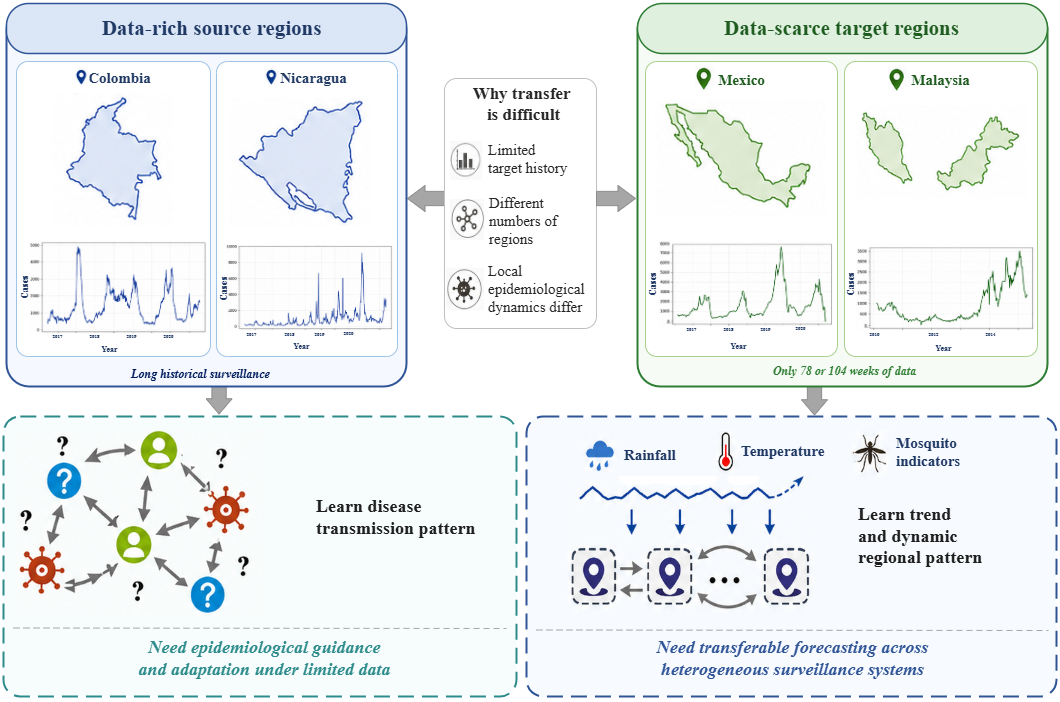}
  \caption{Data-rich surveillance systems (Colombia and Nicaragua) contain long historical dengue records, whereas data-scarce systems (Mexico and Malaysia) provide only 78 or 104 weeks of observations. Differences in surveillance history, epidemiological dynamics, and the numbers of monitored regions make direct transfer difficult, motivating a transferable epidemiologically informed forecasting framework (8-week-ahead dengue forecasting under limited target data).}
  \label{fig:motivation}
\end{figure}

Existing approaches for epidemic forecasting can be organized into three broad paradigms. The first comprises mechanistic compartmental models, such as susceptible–infectious–recovered (SIR) models \citep{kermack1927contribution} and environmental time-series SIR (ETSIR) models \citep{wang2025etsirmsa}. By explicitly representing transmission processes and incorporating epidemiological or environmental knowledge, these models can remain informative when observations are scarce. Their reliance on simplified structural assumptions, however, limits their ability to represent nonlinear interactions, behavioral responses, policy interventions, and other external influences that shape real-world outbreaks \citep{deng2020colagnn, rodriguez2023einns}.

The second paradigm uses data-driven learning to infer temporal and spatial dependencies directly from surveillance records. Recurrent, convolutional, graph-based, and transformer architectures can model complex nonlinear epidemic patterns \citep{deng2020colagnn, xie2022epignn, kamarthi2023profhit, panja2023ensemble}, but typically require long training histories. Many spatiotemporal models are also tied to a fixed graph or number of spatial units, making transfer to surveillance systems with different administrative structures difficult without architectural modification and substantial retraining \citep{rodriguez2021calinet}. More recently, large-scale time-series foundation models have achieved strong zero-shot performance through pretraining on massive cross-domain datasets \citep{panja2025zeroshot, das2024timesfm, ansari2024chronos, auer2025tirex}. Although they reduce the need for target-domain training, their forecasts are driven primarily by statistical patterns learned during pretraining and do not explicitly incorporate local epidemiological mechanisms.

The third paradigm combines the flexibility of neural forecasting with the structural knowledge encoded by compartmental models. Recent hybrid approaches, including Epidemiologically-Informed Neural Networks (EINNs) \citep{rodriguez2023einns}, Epidemic-Guided Deep Learning (EGDL) \citep{barman2025epidemic}, EARTH \citep{wan2025earth}, and DINN \citep{cao2026dinn}, incorporate epidemiological information through mechanistic features, latent disease dynamics, architectural constraints, or physics-informed objectives \citep{rodriguez2022hybrid}. Although these methods can improve forecasting accuracy and epidemiological consistency, they are typically trained end-to-end within a single surveillance system. Consequently, the learned neural representations may become tightly coupled to local transmission parameters, spatial structure, and data availability, making them difficult to transfer to systems with different numbers of monitored regions or only short observational histories. Such coupling can also complicate optimization in data-scarce settings and limits straightforward integration with frozen, zero-shot time-series foundation models. 

This raises the central problem considered in this work (also see Figure~\ref{fig:motivation} for an illustration): {\it How can mechanistic epidemiological knowledge learned from data-rich source regions be transferred to a structurally different, data-scarce target surveillance system while preserving a pretrained forecasting backbone and requiring only minimal target-specific adaptation}?

To address this gap, we introduce TREA-Net (Transferable Residual Epidemiological Adaptation Network), a modular transfer-learning framework for dengue forecasting under limited target data. TREA-Net combines ETSIR-based mechanistic guidance with predictions from any pretrained forecasting backbone through a lightweight residual adapter. The adapter is learned from data-rich surveillance systems and transferred to data-scarce targets without modifying the backbone or assuming a fixed number of monitored regions. This provides a practical approach to portable multi-week dengue early warning for health agencies with limited data and computational resources. Our main contributions are:
\begin{itemize}
    \item {\bf Transferable mechanistic adaptation.} We formulate epidemiological guidance as a latent residual correction using ETSIR projections and pretrained forecasts, rather than embedding mechanistic equations in the training loss. The point-wise, N-invariant adapter contains no node-specific parameters or fixed spatial topology, allowing transfer across surveillance systems with different numbers of administrative units while requiring only two global target-adaptation parameters.
    \item {\bf Multi-country low-data evaluation.} We transfer TREA-Net from Colombia and Nicaragua to Mexico and Malaysia using only 78 or 104 weeks of target observations. Across five forecasting backbones and ten transfer settings spanning 7–33 subnational units, TREA-Net improves the corresponding backbone in nine settings. We further combine it with conformal prediction to provide calibrated uncertainty, with improvements assessed through statistical significance testing.
    \item {\bf Open framework for public-health forecasting.} We will release the TREA-Net implementation and evaluation pipeline to support reproducible assessment of user-defined forecasting models under realistic low-data transfer settings. By reducing data, retraining, and computational requirements, the framework can support earlier vector-control planning, healthcare resource allocation, and risk-informed decision-making in resource-constrained regions.

\end{itemize}

% ============================================================
\section{Related Work and Problem Formulation}

\paragraph{Epidemic Forecasting and Mechanistic–Neural Models.}
Classical compartmental models, including SIR, SEIR, TSIR, and their
environmental extensions, encode disease-transmission mechanisms and relevant covariates but require setting-specific calibration and rely on simplified assumptions \citep{kermack1927contribution,bjornstad2002dynamics,
wang2025etsirmsa}. Data-driven approaches instead learn complex spatial and temporal dependencies from surveillance records. Cola-GNN and EpiGNN model inter-regional transmission through learned graph structures \citep{deng2020colagnn,xie2022epignn}, while PROFHiT imposes probabilistic coherence across hierarchical time series \citep{kamarthi2023profhit}. These models are generally trained and evaluated on a fixed set of regions and do not explicitly address transfer between surveillance systems with different spatial dimensions. 

Hybrid approaches seek to combine both paradigms. EINNs and DINNs introduce compartmental dynamics through physics-informed learning
\citep{rodriguez2023einns,cao2026dinn}, whereas PISID integrates an SIR module with region-specific spatial embeddings \citep{fujita2025pisid}. CALI-Net transfers representations from a historical influenza forecaster to a COVID-affected setting but requires target-specific adaptation \citep{rodriguez2021calinet}. Although these methods improve epidemiological consistency or cross-setting adaptation, their learned components remain coupled to particular diseases, spatial identities, or target-specific
optimization. They therefore do not provide a backbone-agnostic mechanism for transferring a frozen epidemiological correction across surveillance systems with different numbers of regions or for augmenting zero-shot foundation models using limited target data.

\paragraph{Parameter-efficient adaptation and $N$-invariant time series.}
Feature-wise Linear Modulation (FiLM) conditions intermediate representations through learned scale and shift parameters \((\gamma,\beta)\) \citep{perez2018film}, while PETSA adapts frozen time-series forecasters using parameter-efficient test-time updates \citep{medeiros2025petsa}. Related work on variable-dimensional time series includes CPiRi, which promotes channel-permutation-invariant transfer through inter-channel attention and channel-shuffling regularization \citep{xu2026cpiri}, and channel-independent
architectures such as PatchTST, which apply shared temporal weights across channels \citep{nie2023patchtst}. This principle also underlies pretrained time-series models including TimesFM, Chronos, and TiRex \citep{das2024timesfm,ansari2024chronos,auer2025tirex}, whose zero-shot performance on epidemic data has recently been examined \citep{panja2025zeroshot}. These methods establish parameter-efficient adaptation and invariance to channel count, but do not explicitly combine mechanistic epidemic projections with forecasts from an independently trained backbone. TREA-Net builds on this shared-weight principle by learning a point-wise epidemiological correction that transfers across different numbers
of regions and requires only a two-parameter target adapter.

\paragraph{Problem Formulation.}
We consider cross-geographical dengue forecasting from data-rich source surveillance systems to a data-scarce target system. Let the source and target systems contain $N_s$ and $N_t$ subnational regions, respectively, where $N_s$ and $N_t$ need not be equal. For each region, the model receives a 26 -week window of dengue incidence and associated covariates. The target system provides only $K \in\{78,104\}$ weeks of historical observations, corresponding to approximately 1.5 or 2 years of surveillance.

Given forecasts from a pretrained neural backbone and mechanistic projections from an ETSIR model, the objective is to predict dengue incidence for the next $H=8$ weeks in every target region. We seek to learn an epidemiologically informed correction from the source systems that can be transferred to the target without retraining or modifying the forecasting backbone. The transferable component must therefore remain applicable across surveillance systems with different numbers of regions and require only minimal estimation from the limited target history.

\begin{figure*}[!t]
  \centering
  \includegraphics[width=0.9\linewidth]{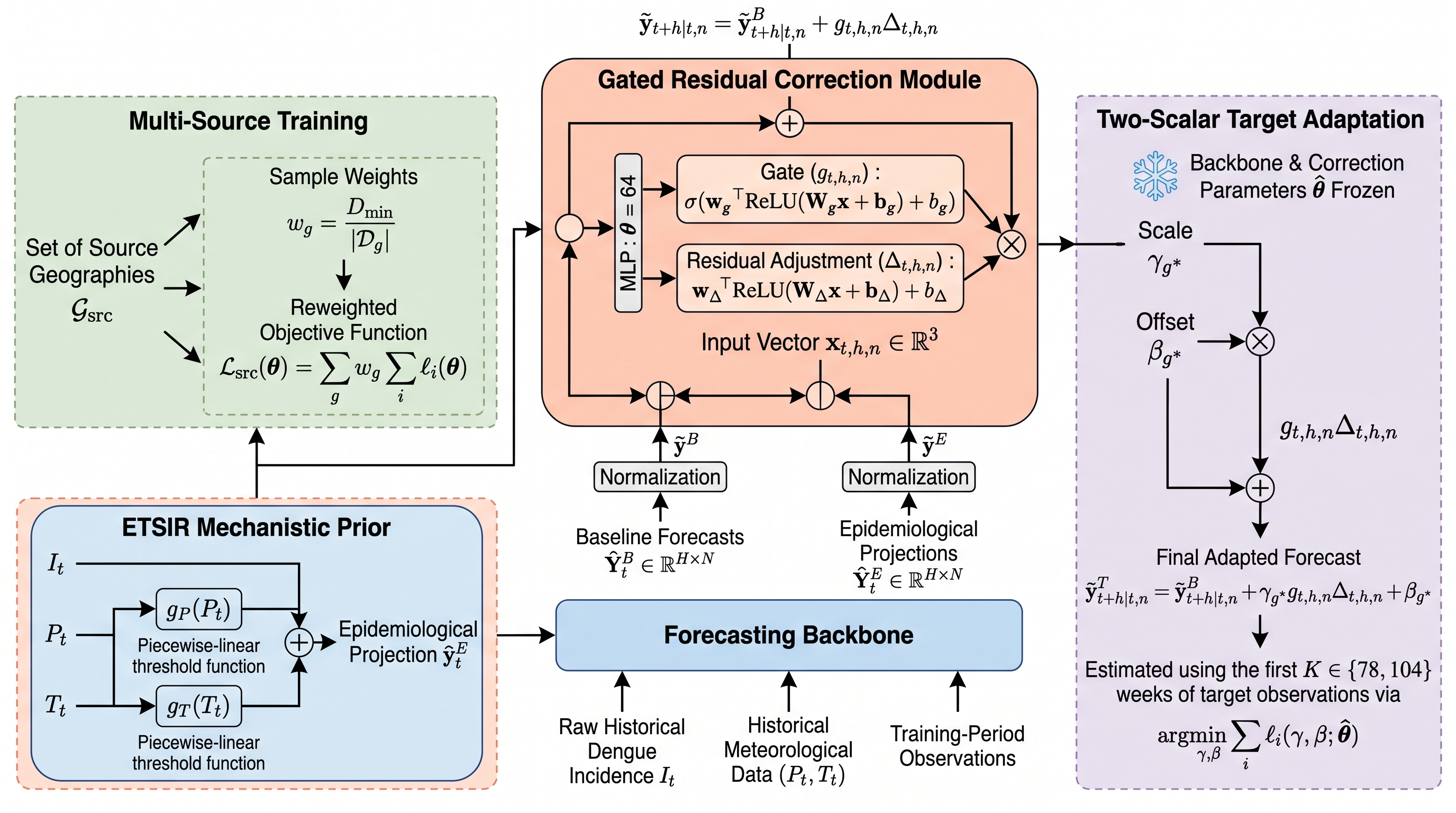}
  \caption{Overview of the proposed TREA-Net framework. ETSIR provides a mechanistic prior; a shared $N$ - invariant gated residual module learns transferable epidemiological corrections from multiple source geographies, and only two target-specific adaptation parameters ($\gamma, \beta$) are estimated for deployment in a new surveillance system.}
  \label{fig:pipeline}
\end{figure*}

\section{Methodology}

\subsection{ETSIR Mechanistic Prior}
To encode climate-sensitive dengue dynamics, we adopt the Environmental Time-Series SIR (ETSIR) model \citep{wang2025etsirmsa}. ETSIR extends classical TSIR dynamics \citep{bjornstad2002dynamics} by combining recent incidence with nonlinear effects of precipitation and temperature. Let $I_t$ denote dengue incidence at week $t$, and define the variance-stabilized response $y_t=\log \left(1+I_t\right)$. Suppressing the regional index for clarity, the ETSIR projection is
\vspace{-0.2em}
\begin{equation}
\widehat{y}_t^E=\left[c_0+c_I \sum_{\ell=1}^L y_{t-\ell}+g_P\left(P_t\right)+g_T\left(T_t\right)\right] y_{t-1},
\label{eq:etsir}
\end{equation}
\noindent where $L$ is the historical lag length and
$$
\begin{aligned}
g_T\left(T_t\right) & =c_T T_t+c_T^{\prime}\left(T_t-H_T\right)_{+}, \\
g_P\left(P_t\right) & =c_P P_t+c_P^{\prime}\left(P_t-H_P\right)_{+},
\end{aligned} \quad(x)_{+}=\max (x, 0) .
$$
The piecewise-linear functions allow the effects of temperature and precipitation to change beyond thresholds $H_T$ and $H_P$, reflecting the non-monotone influence of weather on mosquito abundance and transmission. The lagged-incidence term summarizes recent epidemic activity, including transmission persistence and susceptible-depletion effects. All ETSIR parameters are estimated using training-period observations only. We use multiple-step-ahead (MSA) estimation, which minimizes cumulative forecast error over a multi-week horizon rather than optimizing only one-step predictions. For parameter vector $\theta$, the objective is
$$
\hat{\theta}=\arg \min _\theta \sum_t \sum_{h=1}^{H_{\mathrm{MSA}}}\left(y_{t+h}-\widehat{y}_{t+h \mid t}^E(\theta)\right)^2 .
$$
At test time, future temperature and precipitation are replaced by week-specific climatological averages computed exclusively from the training period, preventing information leakage. The resulting eight-week ETSIR trajectory is not used as a standalone forecast; instead, it provides TREA-Net with a compact mechanistic prior that complements the prediction from the neural backbone.

\subsection{$N$-Invariant Gated Residual Correction}
At forecast origin $t$, let
$\widehat{\mathbf{Y}}_t^B, \widehat{\mathbf{Y}}_t^E \in \mathbb{R}^{H \times N}$
denote the backbone (any forecasting framework) and ETSIR forecasts, respectively, where $H$ is the forecast horizon and $N$ is the number of regions. Their $(h, n)$-th entries are written as $\widehat{y}_{t+h \mid t, n}^B$ and $\widehat{y}_{t+h \mid t, n^{\prime}}^E$ for $h=1, \ldots, H$ and $n=$ $1, \ldots, N$.

After MinMax normalization using training-period statistics, let $\tilde{y}_{t+h \mid t, n}^B$ and $\tilde{y}_{t+h \mid t, n}^E$ denote the normalized forecasts. For each horizon-region pair, we construct
$\mathbf{x}_{t, h, n}=\left[\tilde{y}_{t+h \mid t, n}^B, \tilde{y}_{t+h \mid t, n}^E, \left|\tilde{y}_{t+h \mid t, n}^B-\tilde{y}_{t+h \mid t, n}^E\right|\right]^{T} \in \mathbb{R}^3.$

The correction module estimates a gate $g_{t, h, n}$ and a residual adjustment $\Delta_{t, h, n}$ :
$$
\begin{aligned}
g_{t, h, n} & =\sigma\left(\mathbf{w}_g^{\top} \operatorname{ReLU}\left(\mathbf{W}_g \mathbf{x}_{t, h, n}+\mathbf{b}_g\right)+b_g\right), \\
\Delta_{t, h, n} & =\mathbf{w}_{\Delta}^{\top} \operatorname{ReLU}\left(\mathbf{W}_{\Delta} \mathbf{x}_{t, h, n}+\mathbf{b}_{\Delta}\right)+b_{\Delta}, \\
\tilde{y}_{t+h \mid t, n} & =\tilde{y}_{t+h \mid t, n}^B+g_{t, h, n} \Delta_{t, h, n} .
\end{aligned}
$$
Both multilayer perceptrons use a hidden dimension of 64. The gate $g_{t, h, n} \in(0,1)$ controls the contribution of the residual correction, while $\Delta_{t, h, n}$ determines its direction and magnitude. The corrected forecasts are subsequently mapped back to the original scale using the corresponding inverse transformations. The same functions are applied independently to every $(h, n)$ pair, with no horizon-specific, node-specific, or topology-dependent parameters. The module is therefore invariant to $N$ and can be transferred across surveillance systems containing different numbers of regions. Moreover, the ETSIR projection enters as an input feature rather than through a mechanistic loss term, preserving the forecasting backbone and isolating the transferable epidemiological correction.

\subsection{Multi-Source Training}
Let $\mathcal{G}_{\mathrm{src}}$ denote the set of source geographies, and let $\mathcal{D}_g$ contain the correction-training samples from source $g$. Because the source datasets differ substantially in size, particularly, Colombia contributes approximately 2.4 times as many training windows as Nicaragua; naively pooling all samples would bias learning toward the larger source. We therefore define
$$
D_{\min }=\min _{g \in \mathcal{G}_{\mathrm{src}}}\left|\mathcal{D}_g\right|, \quad w_g=\frac{D_{\min }}{\left|\mathcal{D}_g\right|},
$$
and train the correction parameters $\boldsymbol{\theta}$ using the reweighted objective
$$
\mathcal{L}_{\mathrm{src}}(\boldsymbol{\theta})=\sum_{g \in \mathcal{G}_{\mathrm{src}}} w_g \sum_{i \in \mathcal{D}_g} \ell_i(\boldsymbol{\theta}),
$$
where $\ell_i$ denotes the forecasting loss for sample $i$. This weighting equalizes the aggregate contribution of each source geography without discarding observations. For every source, training samples are generated from all five forecasting backbones and pooled before optimizing the shared correction module. No backbone identifier is supplied to the module; it observes only the normalized backbone forecast, the ETSIR projection, and their absolute discrepancy. Consequently, the learned correction is encouraged to capture relationships that generalize across both source geographies and backbone architectures.

\subsection{Two-Scalar Target Adaptation}
During deployment in a target geography $g^{\star}$, both the forecasting backbone and the source-trained correction parameters $\widehat{\boldsymbol{\theta}}$ remain frozen. We introduce two target-specific global scalars, $\gamma_{g^{\star}}$ and $\beta_{g^{\star}}$, which adjust the magnitude and offset of the transferred correction. For each forecast horizon $h$ and target region $n$,
$$
\tilde{y}_{t+h \mid t, n}^T=\tilde{y}_{t+h \mid t, n}^B+\gamma_{g^{\star}} g_{t, h, n} \Delta_{t, h, n}+\beta_{g^{\star}},
$$
where $g_{t, h, n}$ and $\Delta_{t, h, n}$ are produced by the frozen gated residual module. The two adaptation parameters are estimated using forecasting windows constructed from the first $K \in\{78,104\}$ weeks of target observations:
$$
\left(\widehat{\gamma}_{g^{\star}}, \widehat{\beta}_{g^{\star}}\right)=\arg \min _{\gamma, \beta} \sum_{i \in \mathcal{D}_{g^{\star}}^{\text {adapt }}} \ell_i(\gamma, \beta ; \widehat{\boldsymbol{\theta}}) .
$$
Because $\gamma_{g^{\star}}$ and $\beta_{g^{\star}}$ are shared across all horizons and regions, target adaptation introduces only two trainable parameters, irrespective of the number of administrative units $N_{g^*}$. The corrected forecasts are then mapped back to the original incidence scale. As examined in Section~\ref{sec:ablations}, replacing the global adapter with node-specific scale and shift parameters introduces $2 N_{g^{\star}}$ parameters but provides no consistent improvement and is more susceptible to overfitting under limited target data.

% ============================================================
\begin{table*}[!t]
\centering
\caption{MAE on raw case counts for Malaysia and Mexico under $K \in\{78,104\}$ weeks of target history, averaged over five seeds. \textbf{Bold} and \underline{Underlined} values denote the best and second-best results in each column, respectively. Avg. $\Delta \%$ is the mean percentage reduction in MAE achieved by each TREA-Net variant relative to its paired backbone across the four target-history settings and is reported only for TREA-Net rows. All TREA-Net variants use a single backbone-agnostic correction module trained on reweighted Colombia and Nicaragua source data, with only the target-specific scalars $(\gamma, \beta)$ estimated locally. $\dagger$TiRex is applied zero-shot and produces deterministic backbone forecasts; hence, its standard deviation is zero.}  
\label{tab:main}
\small
\setlength{\tabcolsep}{5pt}
\begin{tabular}{ll rrrr r}
\toprule
 & & \multicolumn{2}{c}{Malaysia} & \multicolumn{2}{c}{Mexico} & \\
\cmidrule(lr){3-4}\cmidrule(lr){5-6}
Tier & Model & $K{=}78$ & $K{=}104$ & $K{=}78$ & $K{=}104$ & Avg\,$\Delta$\% \\
\midrule
\multirow{2}{*}{Reference}
 & ETSIR alone   & 1334.9 & 557.2 & 72.0 & 66.7 & \\
 & ARIMA         &   75.7 &  83.4 & 89.0 & 96.7 & \\
\midrule
\multirow{1}{*}{Physics-informed}
 & ETSIR-PINN         & 48.1 & 49.4 & 68.9 & 75.4 & \\
\midrule
\multirow{5}{*}{Backbone}
 & LSTM      &   34.7 &   43.8 &   71.3 &   65.3 & \\
 & NHiTS     &   28.4 &   29.0 &   68.7 &   61.0 & \\
 & TCN       &   36.6 &   40.0 &   73.6 &   66.4 & \\
 & PatchTST  &   49.3 &   50.5 &   89.5 &   87.7 & \\
 & TiRex$\dagger$ & \underline{23.8} & \underline{26.6} & \underline{54.9} & 60.1 & \\
\midrule
\multirow{5}{*}{{\bf TREA-Net (Proposed)}}
 & TREA-Net-LSTM     &   34.4 &   43.4 &   68.9 &   62.7 &  2.4 \\
 & TREA-Net-NHiTS    &   28.3 &   29.1 &   65.5 & \underline{59.4} &  1.8 \\
 & TREA-Net-TCN      &   36.3 &   39.9 &   65.9 &   64.8 &  3.4 \\
 & TREA-Net-PatchTST &   47.8 &   49.7 &   74.4 &   87.7 &  5.3 \\
 & TREA-Net-TiRex    & \textbf{22.6} & \textbf{25.5} & \textbf{52.4} & \textbf{51.2} & \textbf{7.1} \\
\bottomrule
\end{tabular}
\end{table*}

\section{Experiments}

\subsection{Datasets}
We use weekly subnational dengue surveillance data from four countries. The source pool comprises Colombia, with $N_{\mathrm{Col}}=33$ departments observed over 835 weeks, and Nicaragua, with $N_{\mathrm{Nic}}=18$ departments over 981 weeks. These records span approximately 16 and 19 years, respectively, and represent mature surveillance systems. The target datasets comprise Malaysia, with $N_{\mathrm{Mal}}=15$ states and 270 available weeks, and Mexico, with $N_{\mathrm{Mex}}=22$ states and 208 available weeks. To emulate newly established surveillance programs, target adaptation is restricted to the first $K=104$ weeks of each target series. Weekly temperature and precipitation covariates, together with their 14-, 28-, and 42-day rolling summaries, are aligned with dengue incidence for each region.

We report mean absolute error (MAE) on raw dengue case counts after applying the inverse MinMax transformation. MAE is used because its linear scale provides a direct interpretation in terms of weekly case-count error, making it suitable for downstream resource-allocation analysis discussed in the Real-World Impact section.

\subsection{Forecasting Backbones}
We evaluate five forecasting backbones: LSTM, N-HiTS, TCN, PatchTST \citep{nie2023patchtst}, and TiRex \citep{auer2025tirex}. The first four models use a 26-week input window and a hidden dimension of 64. For backbone-only baselines, these models are trained using only the available target history. To generate source samples for learning the transferable correction, separate backbone instances are trained on the source surveillance systems. TiRex is a pretrained time-series foundation model and is applied in a zero-shot setting without source- or target-specific fine-tuning. Under fixed preprocessing and inference settings, its forecasts are deterministic; hence, its backbone-only MAE has zero variation across repeated runs. For all backbones, TREA-Net operates only on their forecasts and does not modify their internal architectures or parameters.

\subsection{Target Adaptation and Evaluation Protocol}
For each target geography, we consider limited-history regimes of $K \in\{78,104\}$ weeks. The first $K-20$ weeks are used to estimate the two target-adaptation parameters, while the subsequent 20 weeks are used for validation and early stopping. With a 26-week input window and an 8-week forecast horizon, the two regimes yield 25 and 51 adaptation-training windows, respectively. All observations after week $K$ are held out for final testing and are not used for model selection, scaling, or adaptation. The target adapter is optimized using Adam. Learning rates, batch sizes, and early-stopping patience are reported explicitly in the Supplementary Material.

% ============================================================
\subsection{Results and Comparisons}
Table~\ref{tab:main} reports MAE for Malaysia and Mexico under $K \in\{78,104\}$ weeks of target history. Across the five backbones and two target geographies, TREA-Net yields statistically significant improvements in 9 of 10 backbone-geography comparisons according to the Wilcoxon signed-rank test at $\alpha=0.05$ ($p$ values are $<0.05$).

The standalone ETSIR model produces substantially larger errors, particularly for Malaysia at $K=78$ $(\mathrm{MAE}=1334.9)$, illustrating the difficulty of calibrating a purely mechanistic model from short histories. ARIMA is comparatively competitive in Malaysia but performs less consistently in Mexico. Incorporating the ETSIR residual constraint into an MLP reduces MAE by $12.0 \%$ on average and markedly lowers variability across random seeds. For example, the standard deviation decreases from 7.9 to 2.0 in Malaysia and from 14.2 to 0.4 in Mexico at $K=78$. Nevertheless, ETSIR-PINN does not outperform the strongest sequence-model baselines, suggesting that mechanistic regularization alone does not compensate for limited temporal expressivity.

Among the forecasting backbones, zero-shot TiRex achieves the lowest backbone-only MAE in all four target settings. N-HiTS is the strongest target-trained model in Malaysia, whereas LSTM and TCN exhibit greater variation across seeds. Applying the same backbone-agnostic TREA-Net correction improves 18 of the 20 backbone-target-history combinations. Improvements are larger in Mexico, reaching 16.8\% for PatchTST at $K=78$ and $14.8 \%$ for TiRex at $K=104$, while gains in Malaysia range from $0.2 \%$ to $5.1 \%$.

TREA-Net combined with TiRex achieves the lowest MAE in every target setting, reducing error by 7.1\% on average relative to the zero-shot TiRex baseline. The two exceptions are TREA-Net-N-HiTS in Malaysia at $K=104$, where MAE increases by $0.4 \%$, and TREA-Net-PatchTST in Mexico at $K=104$, where the difference is negligible.

\subsection{Ablation Study}
\label{sec:ablations}
Table~\ref{tab:ablate-arch} compares four configurations using the TiRex backbone at $K=104$, with each ablated variant modifying one component of TREA-Net. Removing the ETSIR feature increases MAE by 7.5\% in Malaysia and 11.5\% in Mexico, demonstrating that the mechanistic projection provides information beyond the backbone forecast. Training the gated correction only on the target data performs 5.1\% worse in Malaysia and is comparable in Mexico, indicating that multi-source training improves transfer without requiring a larger target-specific model. Finally, replacing the two global adaptation parameters with $2 N$ node-specific parameters produces no meaningful improvement in Malaysia and changes performance by less than 1\% in Mexico, supporting the simpler two-scalar adapter.

\begin{table}[h]
\centering
\caption{Component ablation of TREA-Net with the TiRex backbone at $K=104$, averaged over five seeds. Avg. $\Delta \%$ denotes the mean percentage increase in MAE relative to TREA-Net-TiRex across Malaysia and Mexico; positive values indicate worse performance. In-domain correction trains the gated correction solely on the target data without source transfer or subsequent adaptation. Per-node adapter replaces the two global parameters $(\gamma, \beta)$ with node-specific scale-shift pairs $\left(\gamma_n, \beta_n\right)$, introducing $2 N$ target parameters.}
\label{tab:ablate-arch}
\small
\begin{tabular}{l rr r}
\toprule
Method & MAL & MEX & Avg\,$\Delta$\% \\
\midrule
TREA-Net-TiRex (reference) & \textbf{25.5} & \textbf{51.2} & --- \\
\midrule
w/o ETSIR feature        & 27.4 & 57.1 & $+9.5$ \\
In-domain adapter        & 26.8 & 51.0 & $+2.4$ \\
Per-node ($2N$) adapter  & 25.5 & 50.8 & $-0.4$ \\
\bottomrule
\end{tabular}
\end{table}

\subsection{Uncertainty Quantification via Conformal Prediction}
\label{sec:conformal}

We construct 90\% prediction intervals using Ensemble Batch Prediction Intervals (EnbPI) \citep{xu2023conformal}, which updates calibration residuals sequentially and is therefore well suited to evolving epidemic conditions. Under the $K=104$ regime, we compare zero-shot TiRex with TREA-Net-TiRex at nominal miscoverage level $\alpha=0.10$ (refer to Supplementary material).

\begin{figure}[t]
  \centering
  \includegraphics[width=\columnwidth]{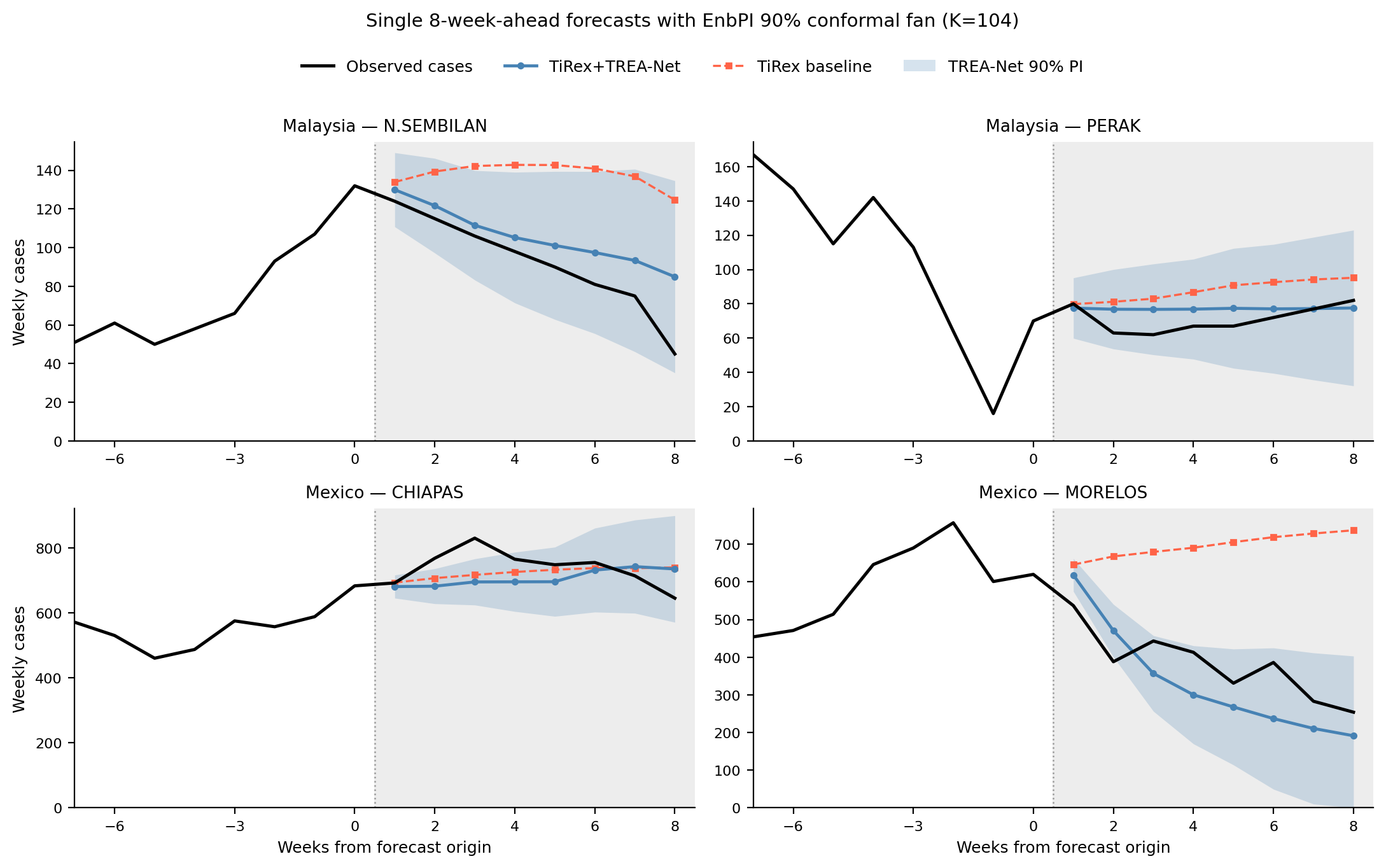}
  \caption{Representative 8-week-ahead forecasts with 90\% EnbPI prediction intervals under K=104. Black lines denote observed weekly cases, with the dotted vertical line marking the forecast origin. Blue lines show TREA-Net–TiRex forecasts and shaded regions their conformal intervals; red dashed lines show the uncorrected TiRex baseline. TREA-Net corrects substantial baseline deviations in Negeri Sembilan and Morelos while preserving accurate forecasts in Perak and Chiapas, with the observed trajectories contained within the prediction intervals.}
  \label{fig:forecast-panels}
\end{figure}

Figure~\ref{fig:forecast-panels} presents representative 8-week-ahead forecasts for four subnational units across Malaysia and Mexico. Each panel shows the observed series, the TiRex baseline, and the TREA-Net-TiRex forecast with its 90\% conformal interval. In Negeri Sembilan and Morelos, TREA-Net corrects baseline trajectories that persist at overly high levels or continue increasing despite an observed decline. In Perak and Chiapas, where TiRex is already accurate, the correction remains small, and the intervals are comparatively narrow. Across the examples, the conformal bands cover nearly all future observations, consistent with the aggregate coverage results.

\subsection{Gate Interpretability}
\label{sec:gate}
\begin{figure}[h]
  \centering
  \includegraphics[width=\columnwidth]{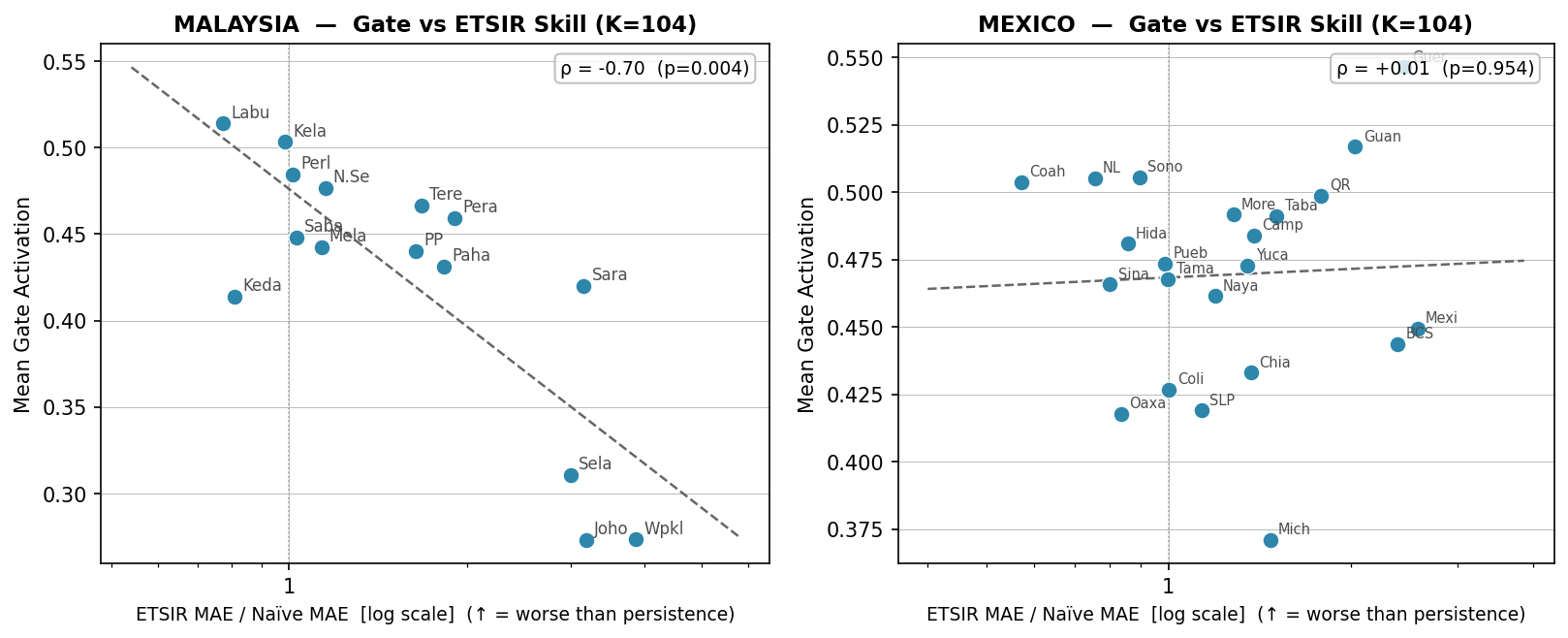}
  \caption{Mean gate activation versus relative ETSIR skill across subnational regions at $K=104$. The horizontal axis reports the ratio of ETSIR MAE to persistence MAE, with values above 1 indicating that ETSIR performs worse than persistence. Gate activation is significantly lower where ETSIR is less reliable in Malaysia ($\rho=$ $-0.70, p=0.004)$, whereas no spatial association is observed in Mexico ($\rho=0.01, p=0.954$).}
  \label{fig:gate-b}
\end{figure}

Figure~\ref{fig:gate-b} compares mean gate activation with relative ETSIR skill across subnational regions. In Malaysia, gate activation decreases as the ETSIR-to-persistence MAE ratio increases ($\rho=-0.70, p=0.004$), indicating that TREA-Net attenuates the transferred correction where the mechanistic prior is less reliable. This spatial association emerges despite the gate being trained only through the forecasting objective, without access to the held-out ETSIR skill metric. No corresponding spatial relationship is observed in Mexico ($\rho=0.01, p=0.954$), where ETSIR performance is more homogeneous across states. Instead, gate activation varies primarily across forecast origins, with a significant negative association with prediction error $(n=97, \rho=-0.267, p=0.008)$; this temporal association is negligible in Malaysia ($n=159, \rho=-0.015, p=0.855$). These results suggest that the gate adapts along the dominant source of predictive variation-spatially in Malaysia and temporally in Mexico.

% ============================================================
% ============================================================
\section{Discussion: Real-World Impact and Limitations}
\label{sec:limitations}
Dengue imposes substantial health and economic costs, exceeding 3 billion USD annually in Latin America and the Caribbean alone \citep{laserna2018economic}. Eight-week forecasts can help health agencies anticipate changes in regional case burden and prioritize vector-control teams, diagnostic capacity, clinical staffing, bed availability, and medical supplies. Forecast improvements should not be interpreted directly as cases averted, since health outcomes depend on subsequent interventions and local operational constraints. Rather, lower case-count errors provide decision-makers with earlier and more reliable information for allocating limited resources.

TREA-Net is designed for surveillance systems that cannot train large forecasting models from scratch. Its transferable correction module contains approximately 5K parameters, while target adaptation estimates only two global scalars and completes on a CPU in under one minute in our implementation. This lightweight design may broaden access to mechanistically informed forecasting capabilities that would otherwise remain concentrated in data-rich and computationally well-resourced settings.

Our analysis uses aggregate weekly case counts at the first subnational administrative level, obtained from public health sources and OpenDengue \citep{clarke2024opendengue}, together with publicly available climate covariates. No patient-level identifiers, addresses, or individual records are processed. Nevertheless, surveillance data may contain under-reporting, changes in case definitions, and spatially uneven ascertainment. TREA-Net should therefore support, rather than replace, public-health judgment. Prospective deployment would require local validation, engagement with health authorities, appropriate institutional and data-governance review, and monitoring for unequal performance across regions.

Several limitations remain. TREA-Net requires $K \in \{78,104\}$ weeks of target surveillance and sufficient climate information to fit a local ETSIR model; it does not address monthly reporting, deployment with no target history, or prolonged reporting disruptions. Future weather inputs are currently approximated using training-period weekly climatology, which cannot anticipate anomalous conditions; operational meteorological forecasts could improve responsiveness to emerging climate-driven risk. Finally, evaluation is retrospective and limited to two target countries. Prospective studies are needed to determine whether improved forecasts lead to better resource allocation, outbreak preparedness, and health outcomes.

% ============================================================
\section{Conclusion}

We introduced TREA-Net, a lightweight transfer-learning framework for dengue forecasting under limited target data. By combining ETSIR projections with backbone forecasts through an N-invariant gated residual correction, TREA-Net transfers epidemiologically informed forecasting knowledge across surveillance systems without node-specific embeddings or backbone fine-tuning. Target adaptation requires estimating only two global parameters, enabling deployment across geographies with different numbers of administrative units. Across Mexico and Malaysia under $K \in \{78,104\}$ weeks of target history, TREA-Net consistently improves diverse neural and foundation-model backbones, with statistically significant gains in 9 of 10 backbone–geography comparisons. The framework also provides calibrated prediction intervals and requires only a small correction module, making it suitable for health agencies with limited surveillance data and computational capacity.

Future work will extend TREA-Net to additional diseases and surveillance systems, incorporate operational weather forecasts instead of historical climatology, and investigate adaptation with shorter or disrupted target histories. Prospective evaluation with public-health partners is also needed to determine how forecast improvements translate into better intervention timing, resource allocation, and outbreak preparedness.

% \section*{Supplementary Material}
% The supplementary material provides complete dataset descriptions, preprocessing procedures, model and optimization settings, computational details, and additional results supporting the reproducibility and robustness of our evaluation.

\bibliography{aaai2026}

\appendix

%\vspace
% \noindent The rest of this document provides supplementary material for the main paper. Appendix~\ref{sec:etsir-why} motivates the use of the Environmental Time-Series SIR (ETSIR) prior, describes the nonlinear effects of climate on dengue transmission, and details the climate-data extraction and preprocessing pipeline. Appendix~\ref{sec:forecast-vis} presents additional qualitative forecasts with conformal prediction intervals. Appendix~\ref{sec:impl} gives implementation details and compute footprint. 

% ============================================================
\section{Environmental Prior: Motivation and Extraction}
\label{sec:etsir-why}

\subsection{Nonlinear Climate Dependence of Dengue Transmission}
Dengue transmission is mediated by \textit{Aedes aegypti}, whose abundance and vectorial capacity respond to temperature and precipitation in nonlinear and often non-monotonic ways. This motivates using an environmentally informed mechanistic prior, rather than representing climate effects only through unconstrained linear covariates.

\paragraph{Temperature.}
Transmission potential is typically unimodal in temperature because biting rate, viral extrinsic incubation, and mosquito survival are each optimized over an intermediate thermal range. Risk increases toward an optimum near $29^\circ\mathrm{C}$, but declines at cooler and hotter extremes, becoming negligible below approximately ${\sim}18^\circ\mathrm{C}$ and above ${\sim}34^\circ\mathrm{C}$ \citep{mordecai2017thermal}. A single linear temperature effect cannot represent this reversal: the same increase in temperature may raise transmission below the optimum but reduce it above the optimum.

\paragraph{Precipitation.}
Rainfall exerts competing effects on mosquito abundance \citep{panja2023ensemble}. Moderate precipitation creates and replenishes standing-water habitats, whereas intense rainfall can flush larvae and eggs from breeding containers. Dengue risk may therefore increase with rainfall up to a threshold and then saturate or decline. The piecewise environmental response in ETSIR provides a parsimonious way to represent such threshold-dependent effects.

\begin{figure}[t]
  \centering
  \includegraphics[width=\columnwidth]{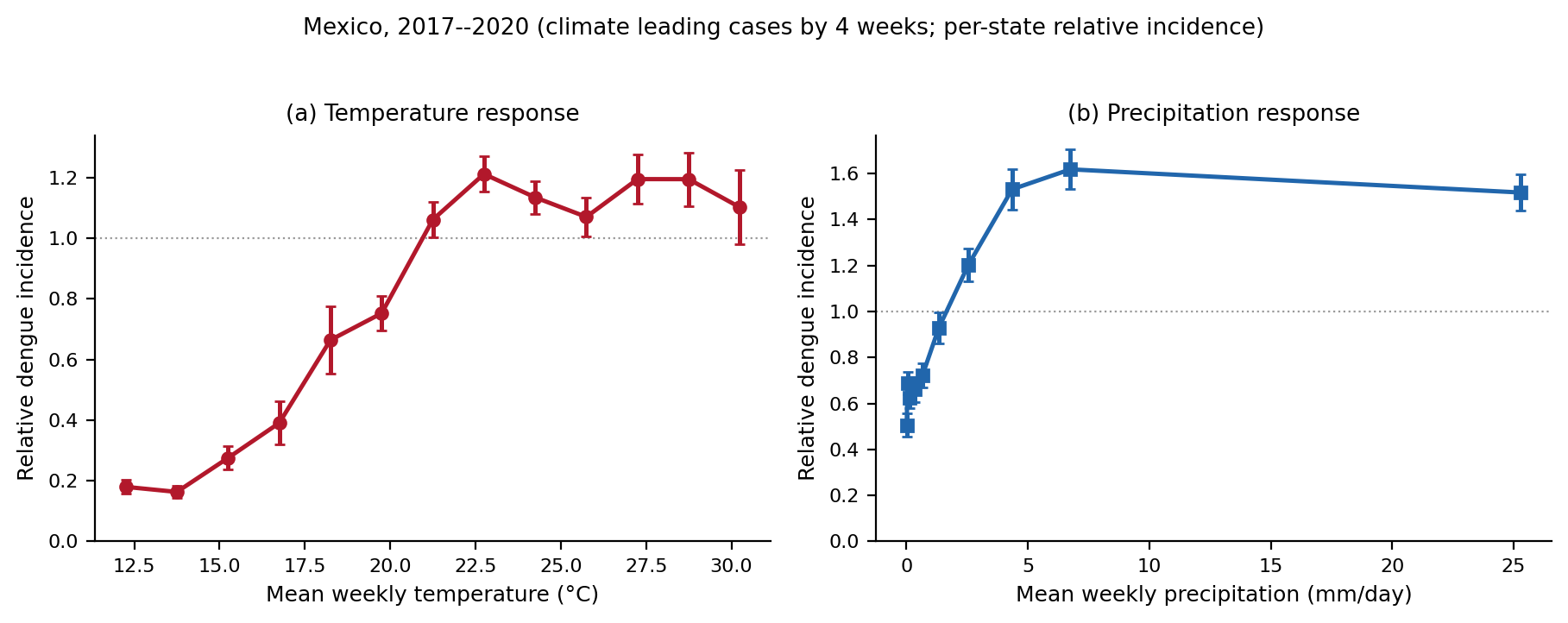}
  \caption{Empirical climate–dengue associations in the Mexico surveillance data (2017–2020). Daily climate series from Google Earth Engine are aggregated weekly and aligned with dengue incidence at a four-week lead. Case counts are normalized within each state so that the curves reflect response shape rather than differences in baseline burden. (a) Incidence increases with temperature to a broad optimum around $23\text{--}29^\circ\mathrm{C}$ and declines at higher temperatures. (b) Incidence rises with moderate rainfall and then saturates. Error bars denote $\pm1$ SEM.}
  \label{fig:climate-nonlin}
\end{figure}

Figure~\ref{fig:climate-nonlin} illustrates these relationships in the target data. Dengue incidence exhibits a hump-shaped association with temperature and a saturating, non-monotonic association with rainfall. These patterns suggest that a single linear covariate effect is inadequate, whereas the breakpoint structure in ETSIR can represent changes in both the magnitude and direction of environmental influence.

\paragraph{Why ETSIR?}
These biological relationships motivate the ETSIR prior used in the main paper. Instead of treating temperature and precipitation as globally linear predictors, ETSIR represents their effects through piecewise-linear forcing functions $g_T(\cdot)$ and $g_P(\cdot)$ with learned breakpoints $H_T$ and $H_P$. The rectified terms ( $T_t-$ $\left.H_T\right)_{+}$and $\left(P_t-H_P\right)_{+}$allow the marginal effects of temperature and rainfall to change beyond these thresholds, thereby accommodating thermal optima, saturation, and potential wash-out effects. The resulting ETSIR projection provides TREA-Net with a compact climate-aware summary of dengue transmission dynamics. Consistent with the ablation results in the main paper, this mechanistic signal contributes information beyond that contained in the backbone forecast alone.

\subsection{Climate Data Extraction}
\label{sec:gee}
The climate covariates used by ETSIR are obtained from publicly available satellite-derived and reanalysis products accessed through Google Earth Engine \citep{gorelick2017gee}. A common extraction and aggregation pipeline is applied across all source and target geographies to ensure consistent and comparable environmental features.

\begin{itemize}
  \item \textbf{Precipitation.}
  Daily precipitation is obtained from the CHIRPS dataset
  (\texttt{UCSB-CHG/CHIRPS/DAILY})~\citep{funk2015chirps},
  which provides quasi-global rainfall estimates at approximately
  \(\sim 5.5\,\mathrm{km}\) spatial resolution.

  \item \textbf{Temperature.}
  Daily \(2\,\mathrm{m}\) air temperature is obtained from ERA5-Land
  (\texttt{ECMWF/ERA5\_LAND/DAILY\_AGGR})~ \citep{munozsabater2021era5land} and converted from Kelvin to degrees   Celsius. Missing pixels along coastal boundaries are filled using a \(3\times 3\)-pixel focal mean before aggregation to the administrative-region level.

  \item \textbf{Vector-viability mask.}
  Both climate products are restricted to elevations below
  \(2000\,\mathrm{m}\) using the SRTM digital elevation model
  (\texttt{USGS/SRTMGL1\_003}). Because \textit{Aedes aegypti}
  occurrence and dengue transmission are substantially reduced at high elevations, this mask prevents cold, high-altitude pixels that are unlikely to support transmission from diluting the
  administrative-region climate summaries.

  \item \textbf{Spatial aggregation.}
  The masked daily rasters are spatially averaged within each first-level administrative polygon obtained from
  \texttt{FAO/GAUL/2015/level1}. The resulting regional climate series are then matched to the department or state names used in the corresponding dengue surveillance data.

  \item \textbf{Temporal aggregation and lagged features.}
  Daily climate series are aggregated to epidemiological weeks, and
  rolling 14-, 28-, and 42-day means are constructed to capture delayed environmental effects on mosquito development and dengue transmission. At operational inference time, climate covariates over the forecast horizon are replaced by week-specific climatological averages estimated exclusively from the training period, as described in the ETSIR Prior section of the main paper.
\end{itemize}

\noindent The extraction pipeline is used solely for data preparation. It operates on publicly available aggregate climate rasters and first-level administrative boundaries and produces one regional table per geography with columns \texttt{(date, department, precip\_mm, temp\_c)}. These tables are subsequently used as inputs to the ETSIR estimation procedure.

% ============================================================

\section{Additional Experimental Details}\label{sec:forecast-vis}

\subsection{Dataset Details}
\label{sec:datasets}
Table~\ref{tab:datasets} summarizes the four subnational dengue surveillance datasets. The source--target assignment reflects surveillance maturity: Colombia and Nicaragua each provide more than 15 years of weekly case records and constitute the transfer-source pool, whereas Mexico and Malaysia are used as targets to emulate newly established surveillance programs. For each target,
only the first \(K\in\{78,104\}\) weeks are available for local model training and adaptation, and all subsequent observations are reserved for out-of-sample testing. Each geography is paired with the CHIRPS and ERA5-Land climate covariates described in Appendix~\ref{sec:gee}, including temperature, precipitation, and their 14-, 28-, and 42-day rolling summaries, aligned weekly
for each administrative unit.

\begin{table}[t]
\centering
\caption{Subnational dengue surveillance datasets. ``Units'' denotes the number of first-level administrative units (departments or states) with usable dengue case and climate series. Target datasets are evaluated under \(K\in\{78,104\}\) weeks of available local history.}
\label{tab:datasets}
\small
\begin{tabular}{llrrl}
\toprule
Country & Role & Units & Weeks & Surveillance span \\
\midrule
Colombia  & Source & 33 & 835 & $\sim$16 years \\
Nicaragua & Source & 18 & 981 & $\sim$19 years \\
Mexico    & Target & 22 & 208 & 2017--2020 \\
Malaysia  & Target & 15 & 270 & 2010--2015 \\
\bottomrule
\end{tabular}
\end{table}

\subsection{Horizon-Wise Prediction-Interval Efficiency}
\begin{figure}[h]
  \centering
  \includegraphics[width=\columnwidth]{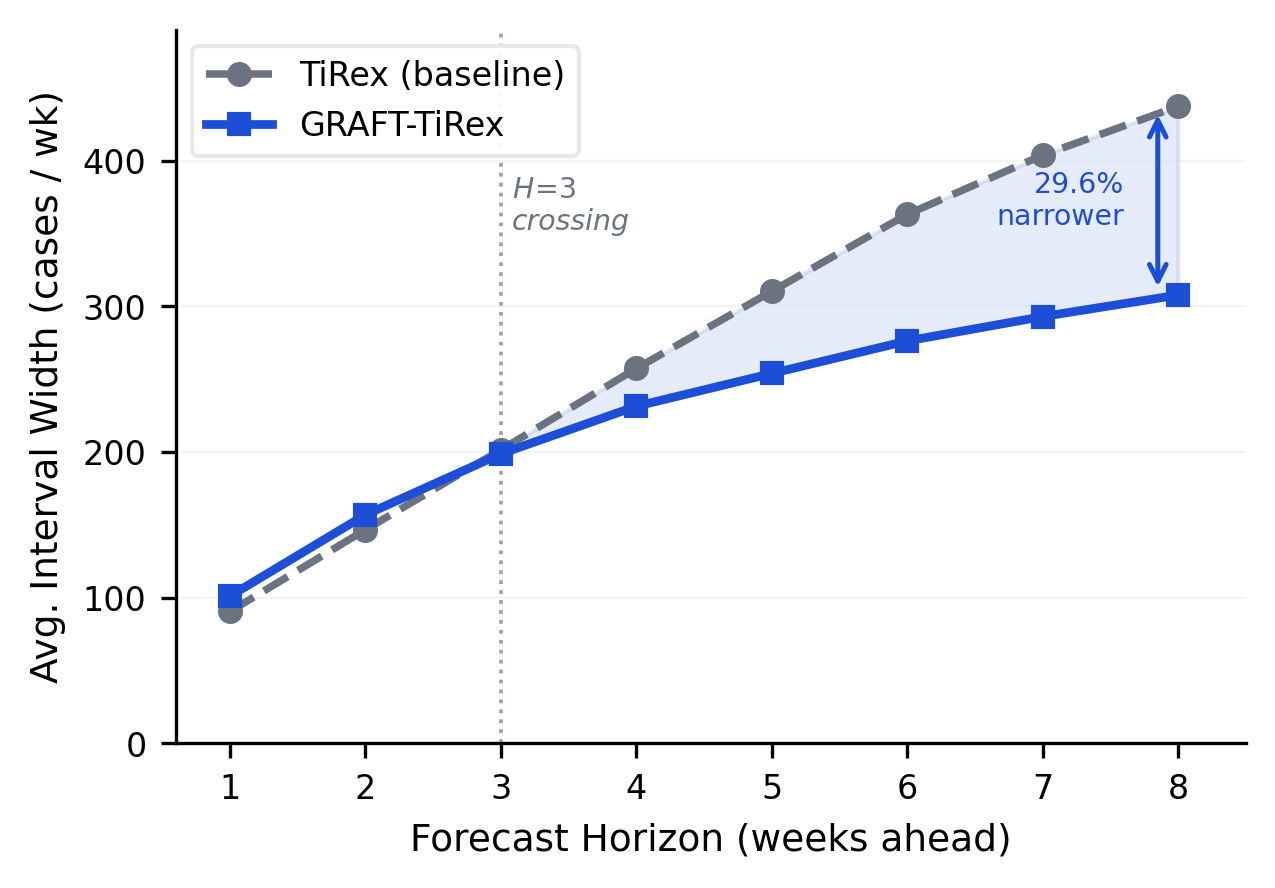}
  \caption{Average EnbPI prediction-interval width by forecast lead in Mexico under \(K=104\) and nominal miscoverage level \(\alpha=0.10\). The shaded region indicates forecast leads for which TREA-Net--TiRex produces narrower intervals than the TiRex baseline. Both methods attain approximately \(87\%\) empirical coverage over the test period.}
  \label{fig:conformal-mexico}
\end{figure}

Although empirical coverage remains broadly stable, TREA-Net changes the width of the conformal prediction intervals, particularly at longer forecast leads. In Mexico, the uncorrected TiRex backbone produces narrower intervals at short leads (\(h\in\{1,2\}\)), consistent with its strong near-term autoregressive
performance. As the forecast lead increases, however, the TiRex intervals widen more rapidly. At \(h=8\), TREA-Net--TiRex produces intervals that are \textbf{29.6\% narrower} than those of TiRex while maintaining comparable empirical coverage (Figure~\ref{fig:conformal-mexico}). This result suggests
that the ETSIR-informed residual correction improves interval efficiency at longer leads by stabilizing the underlying point forecasts. 

In Malaysia, TREA-Net and TiRex produce intervals of comparable width across the forecast horizon, consistent with the smaller corrections observed in this target geography. Overall, these results indicate that incorporating the mechanistic projection can yield tighter prediction intervals at longer forecast leads without an evident loss of empirical coverage.

\subsection{Empirical Coverage under Distribution Shift}
Both methods exhibit some under-coverage relative to the nominal \(90\%\) level, attaining approximately \(87\%\) empirical coverage in Mexico and \(81\%\) in Malaysia. This shortfall coincides with a shift from predominantly low-incidence validation periods to test periods containing larger epidemic peaks, for which forecast residuals are substantially greater. Despite this shift, TREA-Net
preserves the empirical coverage of the corresponding zero-shot TiRex baseline. Thus, the ETSIR-informed correction improves point forecasts and interval efficiency without an observed deterioration in empirical coverage.

\subsection{Statistical Significance Testing}
For each backbone--target pair, we compare TREA-Net with its corresponding backbone across the two target-history regimes,
\(K\in\{78,104\}\), and five random seeds. This yields \(10\) paired MAE differences per backbone--target comparison. We apply a two-sided Wilcoxon signed-rank test to these paired differences at significance level \(\alpha=0.05\). TREA-Net achieves statistically significant improvements in 9 of the 10 backbone--target comparisons.

% ============================================================

% ============================================================
\section{Implementation Details and Compute Footprint}
\label{sec:impl}

Table~\ref{tab:hparams} summarizes the architecture and optimization settings for all trainable components. The correction module and the trainable forecasting backbones are optimized with early stopping using a held-out validation window and a patience of 20 epochs. TiRex is applied zero-shot and is therefore not updated. At deployment, the backbone and the source-trained correction module remain frozen, while only the two target-specific parameters
\((\gamma,\beta)\) are estimated from the adaptation portion of the available local target history.

\subsection{Compute Footprint}
The transferable component is the source-trained correction module, containing approximately \(5{,}000\) parameters, which is trained once on the source pool and subsequently frozen. For a new target geography, only the two global adaptation parameters \((\gamma,\beta)\) are estimated. In our implementation,
this optimization completes in under one minute on a CPU and does not require GPU acceleration. Backbone and ETSIR forecasts are generated once and cached before adaptation. Consequently, the incremental computation required to adapt TREA-Net to a new surveillance system is small relative to retraining or fine-tuning the forecasting backbone.

\begin{table}[t]
\centering
\caption{Architecture and optimization settings for all model components. Maximum epoch counts are upper bounds; early stopping with a patience of 20 epochs typically terminates training earlier. TiRex is applied zero-shot and its parameters were not updated.}
\label{tab:hparams}
\small
\begin{tabular}{ll}
\toprule
Setting & Value \\
\midrule
Input sequence length (IL) & 26 weeks \\
Forecast horizon ($H$)     & 8 weeks \\
Validation window          & 20 weeks \\
Seeds                      & 1--5 \\
\midrule
\multicolumn{2}{l}{\emph{Correction module (Gate+Delta, $N$-invariant)}} \\
\quad hidden dimension        & 64 \\
\quad dropout / gate-bias init & 0.5 / 0.2 \\
\quad parameters              & $\sim$5K \\
\quad optimizer / learning rate & Adam / $5{\times}10^{-5}$ \\
\quad max epochs / patience   & 300 / 20 \\
\quad batch size              & 32 \\
\midrule
\multicolumn{2}{l}{\emph{Neural backbones (LSTM, NHiTS, TCN, PatchTST)}} \\
\quad hidden dimension        & 64 \\
\quad optimizer / learning rate & Adam / $10^{-3}$ \\
\quad max epochs / patience   & 300 / 20 \\
TiRex backbone                & zero-shot (untrained) \\
\midrule
\multicolumn{2}{l}{\emph{$(\gamma,\beta)$ target adapter}} \\
\quad trainable parameters    & 2 scalars \\
\quad optimizer / learning rate & Adam / $10^{-2}$ \\
\quad max epochs / patience   & 200 / 20 \\
\quad fit cost                & $<1$ min, CPU \\
\midrule
Conformal intervals (EnbPI)   & $\alpha=0.10$ (90\% nominal) \\
\bottomrule
\end{tabular}
\end{table}

\end{document}